\documentclass{article}
\usepackage{smc2024}
\usepackage[caption=false, font=footnotesize]{subfig}
\usepackage{paralist}
\usepackage[figure,table]{hypcap}
\usepackage[none]{hyphenat}

\usepackage[english]{babel}

\def\papertitle{From MIDI to rich tablatures: an automatic generative system incorporating lead guitarists' fingering and stylistic choices}

\author[1]{\mbox{\firstname{Pierluigi}\lastname{Bontempi}\orcid{0000-0002-4360-8708}}}
\author[2]{\mbox{\firstname{Daniele}\lastname{Manerba}\orcid{0000-0002-3502-5289}}}
\author[3]{\mbox{\firstname{Alexandre}\lastname{D'Hooge}\orcid{0000-0003-1634-3406}}}
\author[4]{\mbox{\firstname{Sergio}\lastname{Canazza}\orcid{0000-0001-7083-4615}}}

\affil[1,4]{\department{Centro di Sonologia Computazionale, Department of Information Engineering}\institution{University of Padua}\country{Italy}\affiliationtype{University}}
\affil[2]{\department{Department of Information Engineering}\institution{Università degli Studi di Brescia}\country{Italy}\affiliationtype{University}}
\affil[3]{%
 \institution{Univ. Lille, CNRS, Centrale Lille, UMR 9189 CRIStAL}
 \postcode{F-59000}
 \city{Lille}
 \country{France}
 \affiliationtype{University}}

\completesetup

\title{\papertitle}
\begin{document}
	\capstartfalse
	\maketitle
	\capstarttrue

	\begin{abstract}
Although the automatic identification of the optimal fingering for the performance of melodies on fretted string instruments has already been addressed (at least partially) in the literature, the specific case regarding lead electric guitar requires a dedicated approach. 
We propose a system that can generate, from simple MIDI melodies, tablatures enriched by fingerings, articulations, and expressive techniques. 
The basic fingering is derived by solving a constrained and multi-attribute optimization problem, which derives the best position of the fretting hand, not just the finger used at each moment.
Then, by analyzing statistical data from the \textit{mySongBook} corpus, the most common clichés and biomechanical feasibility, articulations, and expressive techniques are introduced. Finally, the obtained output is converted into \textit{MusicXML} format, which allows for easy visualization and use. 
The quality of the tablatures derived and the high configurability of the proposed approach can have several impacts, in particular in the fields of instrumental teaching, assisted composition and arranging, and computational expressive music performance models.
	\end{abstract}

\section{Introduction}\label{sec:introduction}
Fretted string instruments (which include guitar, banjo, ukulele, and bass guitar, among others) allow the musician to play most of the notes on at least two different strings, at different frets. In addition, distinct fingers (usually index, middle, ring, or little finger) of the fretting hand can be used to produce the same note, on the same string and at the same fret. In some cases the notes can also be played on the open strings. In music, the term \textit{fingering} is used to indicate which fingers and, consequently, hand positions should be used to perform a part \cite{Lindlay_2001_fingering_grove}. Opting for well-reasoned fingerings is crucial, either for expressive reasons or to enable the musician to minimize performance effort.

The performer can then make choices, in addition to fingering, regarding articulations and expressive techniques from a range of options generally associated with the specific instrument being played. The term \textit{articulation} refers to how a single note is sounded. Each articulation corresponds to a specific performance technique and generates sounds endowed with distinctive sonic characteristics. The parameters involved may be attack, decay, timbre, dynamics, and in some cases even pitch. The term 
\textit{expressive technique} instead refers to all those techniques used to imprint variations in sound parameters for specifically expressive purposes. Clearly, there may be at least partial overlap between the two categorizations. Since we specifically address lead electric guitar parts, in this context the most common articulations are those of the plucked string (usually with the use of a plectrum) and the legato obtained through \textit{hammer-on} (a finger of the fretting hand strikes the string from above, causing it to resonate) or \textit{pull-off} (a finger of the fretting hand ``plucks/pulls'' the string, when it is already vibrating, reaching a note lower than the preceding one). In terms of expressive techniques, we consider \textit{vibrato} (the pitch of the note is made to oscillate with the use of the finger(s) of the fretting hand or the whammy bar), \textit{slide legato} (or glissando), and \textit{bending} (pushing the fingered string upward or pulling it downward, perpendicular to the neck, the pitch of the note is raised) \cite{Grimes_2014_guitar_physics}. 

In this paper, we propose an automatic system able to generate, from monophonic MIDI melodies (including note-on and note-off messages only) \cite{Moog_1986_midi}, electric guitar tablatures enriched by fingering, articulations, and expressive techniques. 
The basic fingering is derived by solving an optimization problem that considers the possibility of playing the same notes on different string-fret combinations or on open strings as well as biomechanical constraints, and that tries to minimize the time needed for the hand movements and the performer's discomfort, 
while also accounting for timbre characteristics associated with the generated tablature. 
The system also keeps track of the position of the fretting hand, not just of the finger used at each moment.
Then, taking into account the most common clichés and biomechanical feasibility, articulations, and expressive techniques are introduced.
The usage frequencies of such features are derived by analyzing statistical data from the \textit{mySongBook} corpus, an impressive proprietary collection of guitar-oriented transcriptions, already used in the past in other publications \cite{Cournut_2021_most-used-positions, DHooge_2023_bends}.
Finally, for ease of visualization and use, the result is converted into \textit{MusicXML} format.

Two important observations are in order. 
First, we address by choice guitar solos and melodies composed of single notes only and leave the possibility of playing multiple notes simultaneously for future system evolutions. However, 
it should be noted that lead guitar sections include a very high percentage (\textgreater 90\%) of single notes (see Cournut et al.\ 2021 \cite{Cournut_2021_most-used-positions}). 
Second, our system is based on symbolic encodings of music, i.e., MIDI, a specially developed dataframe, and MusicXML. Clearly, the automatic generation of transcriptions from audio recordings of guitar performances, a subject of significant attention in the literature (see, among other works,\cite{Michelson_2018_Tablature,Wiggins_2019_Tablature}), shares the same object of study, but observes it backward, i.e., from performance to tablature or score.
In our case, the goal is to move from a minimalistic score (represented in MIDI format) to a guitar tablature enriched with fingering, articulations, and expressive techniques.

The contribution of this work is multifaceted.
On the one hand, earlier approaches to automatic fingering generation based on optimization techniques are somehow revived and greatly expanded (see Section \ref{sec:fingering}). 
To the best of our knowledge, none of the approaches for automatic fingering generation presented in the past considered simultaneously all of the parameters and attributes described later, nor allowed for a level of user-side customization comparable to that proposed here.\
On the other hand, the automatic insertion of articulations and expressive techniques is deeply explored, while being a largely marginal subject of research in the past (a notable exception is the work by Sarmento et al.\ 2023 \cite{Sarmento_2023_ShredGP}, although focused on automatic music generation).\ We also believe that this work can constitute an important piece toward the construction of models for computational musical expressiveness, in teaching, composition, assisted arranging of guitar parts, and, indirectly, in the stylistic analysis of guitarists. Lastly, the \textit{mySongBook} corpus confirms itself to be a precious resource for understanding contemporary guitar practice and for the generation of automatic systems.

The rest of the paper is organized as follows. Section \ref{sec:fingering} presents the optimization approach followed to derive the basic fingering, while Section \ref{sec:articulations} presents first the \textit{mySongBook} corpus and the statistical data derived from it, and then describes the rule system developed for the automatic addition of articulations and expressive techniques. In Section \ref{sec:MusicXML}, the generation of the MusicXML code is introduced and an example lead guitar part processed by the system is presented, in the form of \textit{rich} tablature and score in standard notation. Finally, we conclude in Section \ref{sec:conclusions}, also discussing some possible applications and future developments.
		
\section{From a MIDI melody to the basic fingering}
\label{sec:fingering}
The automatic search for valid or optimal fingerings in relation to specific musical instruments has a scientific literature of significant size that cannot be entirely reviewed here.\ Hence, we first report the main existing approaches related to plucked string instruments, highlighting the gaps we are trying to fill. Then, we describe our novel and comprehensive methodology.

\subsection{Existing approaches for fingering optimization}

Sayegh (1989) \cite{Sayegh_1989_fingering_string_instruments} tackles the fingering procedure by solving a Shortest Path Problem (SPP) on a weighted and layered directed graph (a so-called \textit{Viterbi} network) in which each layer corresponds to each note of the sequence and contains nodes corresponding to admissible fingerings of such a note. Since only position and string changes between a note and the consecutive one affect the fingering attributes, the weights of such attributes can be directly associated with arcs between two consecutive layers. Concerning the considered attributes, two broad classes of rules may be identified. The first one relates to ease of execution and the second to the homogeneity of the sound generated. 

A similar approach, albeit slightly more refined, is proposed in Radicioni et al. (2004) \cite{Radicioni_2004_Segmentation}. Here, not only movements along and across the fretboard are taken into consideration, but also finger span, and minimum/maximum spans between each pair of fingers are used to evaluate the level of difficulty of the transitions between notes and identify when hand repositioning is needed. The sequence is segmented into musical phrases at the preprocessing phase and the optimization takes place first of all within the individual phrases, then a valid connection between phrases is created. 
The approach of Miura et al. (2004) \cite{Miura_2004_tab_generation} is also based on the segmentation of the melody while the objective is to minimize the number of movements of the fretting hand wrist (along the neck) within each section. The trade-off emerges between the increase of finger span and the hand position changes, where the latter is however considered less uncomfortable for a beginner. Interestingly enough, techniques such as hammer-on, pull-off, and slide are considered. 

Other published works focus more on algorithmic aspects of the fingering. 
In particular, Tuohy and Potter (2005) \cite{Tuohy_2005_genetic} propose a Genetic Algorithm for creating polyphonic guitar tablatures. Their fitness function is based on hand movements along and across the neck, and fretting hand configurations. However, the system does not include information about which finger should be used to play each note. 
A Hidden Markov Model (HMM) is used in Hori et al.\ (2013) \cite{Hori_2013_HMM} to derive the basic fingering of monophonic and polyphonic guitar parts. The fretboard hand arrangements are considered as the hidden states, and the note sequence as the observed one.
In a subsequent work by Hori and Sagayama (2016) \cite{Hori_2016_minimax}, the authors try to simplify the most difficult movements required to play a part, instead of aiming at minimizing the difficulty of the overall execution.
The paper introduces a variant of the basic algorithm for multi-layered networks, called the \textit{minimax Viterbi algorithm}, to solve the problem. When applied to an HMM-based fingering decision network, the algorithm looks for the sequence of the hidden states that maximizes the minimum transition probability on the sequence (transition probabilities are large for easy moves and small for difficult ones). 
The reader is referred to Roohnavazfar et al. (2019) \cite{ROOHNAVAZFAR2019100124} for more recent and sophisticated approaches for finding optimal paths in general multi-stage decision networks.

Finally, a Dynamic Programming approach is used in Raboanary et al. (2017) \cite{Raboanary_2017_bass} to minimize performance effort on monophonic bass guitar lines, with special emphasis on limiting movements along the neck.

\subsection{A new comprehensive optimization framework}


Our approach for generating the basic fingering consists in the formalization, development, and resolution of an optimization problem, which generalizes all those already presented in the literature that appeared under the simplified form of an optimal path search through a multi-layered network (see, e.g., Sayegh 1989 \cite{Sayegh_1989_fingering_string_instruments}). In particular, we enrich the quality of the fingering by introducing i) additional requirements, that affect the available feasible fingerings for each note of the melody, and ii) simultaneous attributes to optimize, that affect the cost of the transition from the fingering of a note to that of the following one.

Concerning the attributes of the fingering to be optimized, we identified three macro-categories:
\begin{itemize}
    \item \textit{time-related} attributes: relate to movements of the fretting hand along (\textit{position change}, PC) and across (\textit{string change}, SC) the neck. Their weights depend on the time needed to physically reach the fingering by the hand and the fingers. 
    \item \textit{discomfort-related} attributes: relate to finger span (\textit{hand spread}, HS) and to playing notes on specific parts of the fretboard (e.g., very near or very far from the headstock). Their weights depend on the physical discomfort generated by the fingering in the player. 
    \item \textit{timbre-related} attributes: relate to attributes that affect the type of sound output produced. Their weights depend on the sound quality generated by the fingering, according to the player's intent and taste. In this regard, we consider here the frequency of use of open strings.
\end{itemize}
We remark that, differently from most of the existing approaches, the system keeps track of the fretting hand position (i.e., the fret at which the index finger is located at each moment) to estimate the spreads between fingers and the longitudinal hand movements. 

Moreover, our optimization setting provides the user with great flexibility, allowing the customization of several aspects and preferences. In particular, the system can be finely configured about:
\begin{itemize}
    \item the intrinsic characteristics of the instrument (number of strings and frets, and notes generated by each open string or string/fret combination);
    \item the fretboard area on which it is more comfortable to play. Typically the preferred area is the middle of the neck, between the fifth and twelfth positions, as noted by Cournut et al. (2021) \cite{Cournut_2021_most-used-positions} and by Heijink et al. (2002) \cite{Heijink_2002_complexity} concerning classical guitar;
    \item the maximum and the minimum allowed spread between each finger and the index finger. Constantly keeping track of the position of the index finger allows for a potentially more accurate assessment of the hand spread than the approach traditionally followed, which only monitors the spread between the finger used to play one note and the one used for the immediately following one. It also makes it possible to correctly assess position changes, i.e., shifts of the hand along the neck (see below); 
    \item the time required for the hand to make longitudinal movements (proportional to the number of frets to be covered) and vertical movements (proportional to the number of strings covered). Transitions for which the \textit{Inter Onset Interval} (IOI) is not sufficient are discarded from the possible alternatives;
    \item how fingers are used. For example, it could be denied to use the same finger to play two contiguous notes unless they are played at the same fret (as by a \textit{barrè} o \textit{semi-barrè} technique) or on the same string (so to allow sliding techniques);
    \item the proportional level of penalty to be associated with each attribute to be optimized, e.g., for position changes, string changes, use of open strings, finger spread, and distance from the optimal fretboard area.
\end{itemize}


\subsection{Implementation details}

In the following, we briefly describe the system implementation from a purely technical point of view. 

Given a monophonic melody in MIDI format, a dataframe is generated in Python language using the \textit{Mido} library (\url{https://github.com/mido/mido}, last accessed: 2024/03/07). For each note, the dataframe contains pitch, start and end positions in time, duration, IOI, and all the data needed later to generate the final MusicXML file. Data on the structure of the instrument (number of strings and frets, and notes generated by each open string and string/key combination) and the configuration parameters of our optimization problem are also generated.

The obtained dataframe is then processed through the \textit{IBM ILOG CPLEX Optimization Studio} software, v20.1.0 (\url{https://www.ibm.com/products/ilog-cplex-optimization-studio}, last accessed: 2024/03/07), in which we implemented our optimal fingering problem as a static and compact Mathematical Programming formulation. The problem is described via the CPLEX's dedicated \textit{Optimization Programming Language}.\
The software returns the string/fret combination (and the associated finger to be used to press the string) or the open string on which to play each note implying the best achievement in terms of time-, discomfort-, and timbre-related attributes.

\begin{figure*}[!h]
    \centering
    \includegraphics[width=15cm]{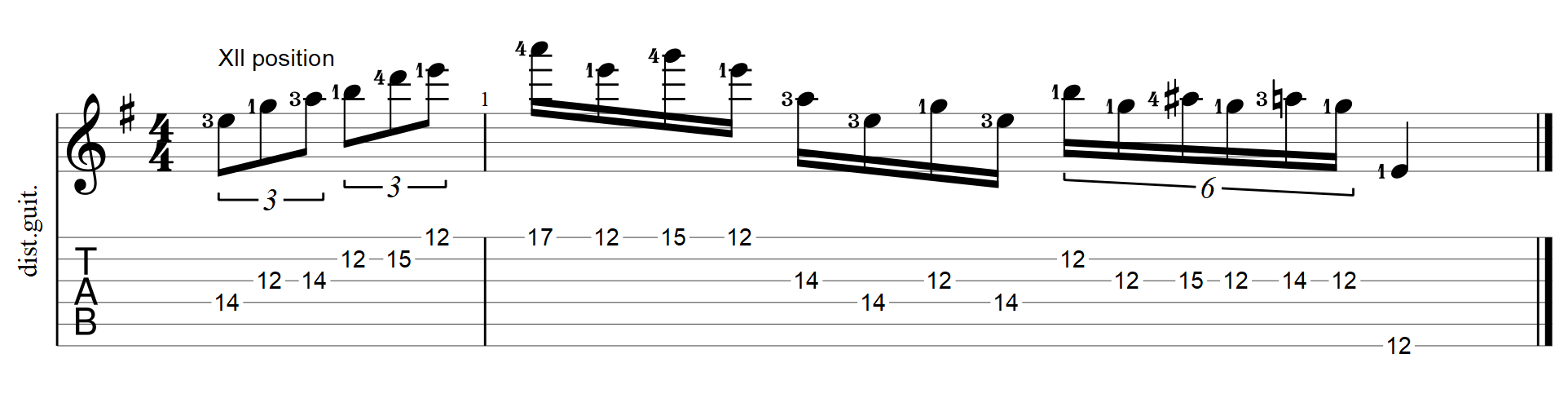}
    \caption{To obtain this solution, the least penalized feature was the string change (SC). The hand spread (HS) was penalized twice as much as SC, while the position change (PC) was penalized twice as much as HS. \label{fig:lick1}}
\end{figure*}
\begin{figure*}[!h]
    \centering
    \includegraphics[width=15cm]{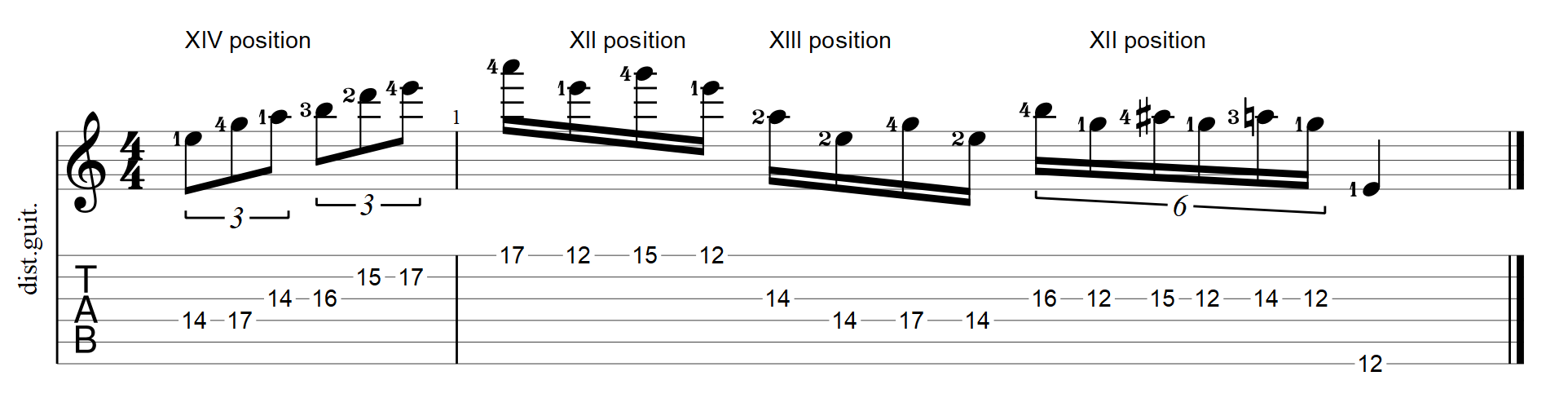}
    \caption{To obtain this solution, the least penalized feature was HS. PC was penalized twice as much as HS. SC was penalized twice as much as PC.  \label{fig:lick2}}
\end{figure*}
Figures \ref{fig:lick1} and \ref{fig:lick2} show the same melody processed by the optimizer with different attributes' importance. The numbers placed next to the heads of the notes in standard notation refer to the finger of the fretting hand to be used, i.e., 1-index, 2-middle, 3-ring, and 4-little finger. Both results are biomechanically valid, but significantly different from each other, reflecting the high configurability of the approach.
	

        \section{From the basic fingering toward a \textit{rich} tablature}\label{sec:articulations}

In this Section, we describe how the dataframe is enriched with articulations and expressive techniques. First, we extracted the distributions of the main techniques from the \textit{mySongBook} dataset
to use them as statistical references. Then, we use a rule-based system for the actual insertion of the techniques, allowing wide freedom of configuration to the user.

\subsection{Statistical data extraction}
\label{sec:stats}
\begin{figure}[ht]
    \centering
    \includegraphics[width=8cm]{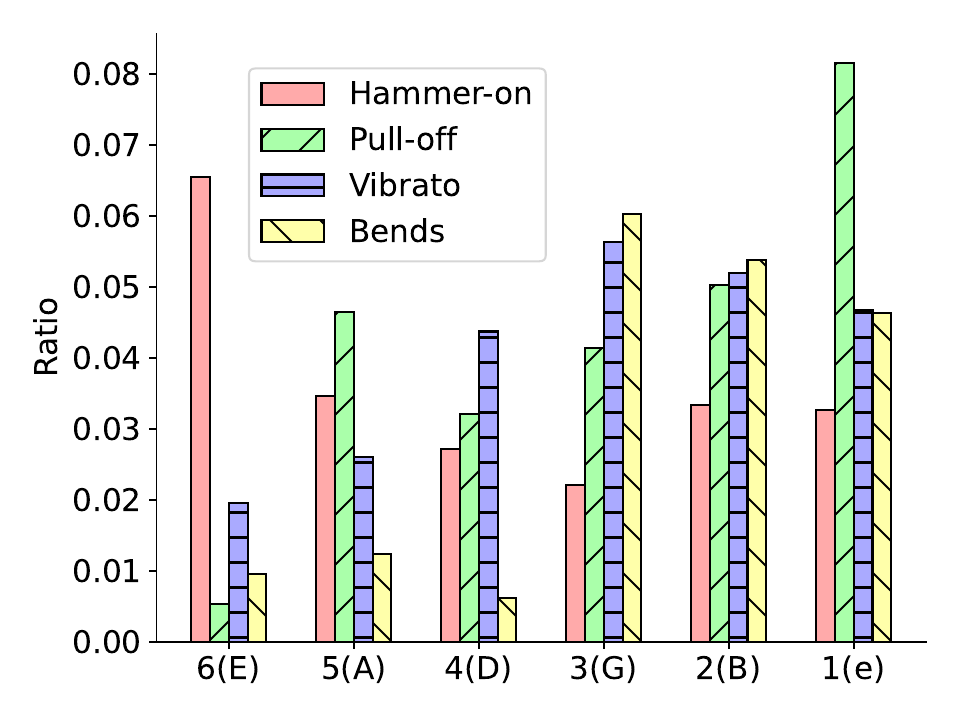}
    \caption{Ratio of the use of the considered techniques to the total number of notes for each string. Strings are labeled according to guitar standard tuning, the 6-th string being the low ``E''
    and the 1-st string the high ``e''.}
    \label{fig:techniques}
\end{figure}

The \textit{mySongBook} corpus comprises more than 2000 scores/tablatures transcribed by professional musicians (\url{https://www.guitar-pro.com/c/18-music-scores-tabs}, last accessed: 2024/03/07), of different genres.\ The transcriptions are made with \textit{Guitar Pro} software, produced by \textit{Arobas Music}, and are geared first and foremost to an audience of guitarists.\ However, they usually also contain vocal parts or parts of other instruments, such as bass guitar or drums.\ Given the object of investigation of this paper, the statistical data described below were extracted from the sections labeled as \textit{lead electric guitar parts}.\ These parts are extracted after eliminating rhythmic guitar sections, identified using the procedure described in Régnier et al.\ (2021) \cite{regnier_identification_2021}.
Then, the quantity of playing techniques on each string is determined and divided by the total number of notes per string to obtain the proportion of notes with each technique. The obtained ratios are illustrated in \autoref{fig:techniques}.

It can be observed that vibrato and bends are used similarly on higher strings. However, bends are rare on lower strings
because thicker strings are harder to bend, while a vibrato is a smaller movement that is manageable even on stiffer strings.
It should also be noted that, even if hammer-on and pull-off are techniques theoretically similar, their distribution across strings is quite different.
Our system uses the statistical data presented for deciding on the insertion of articulations and expressive techniques, as described in Section \ref{sec:articulations-inclusion}. 

\begin{figure}[ht]
    \centering
    \includegraphics[width=8cm]{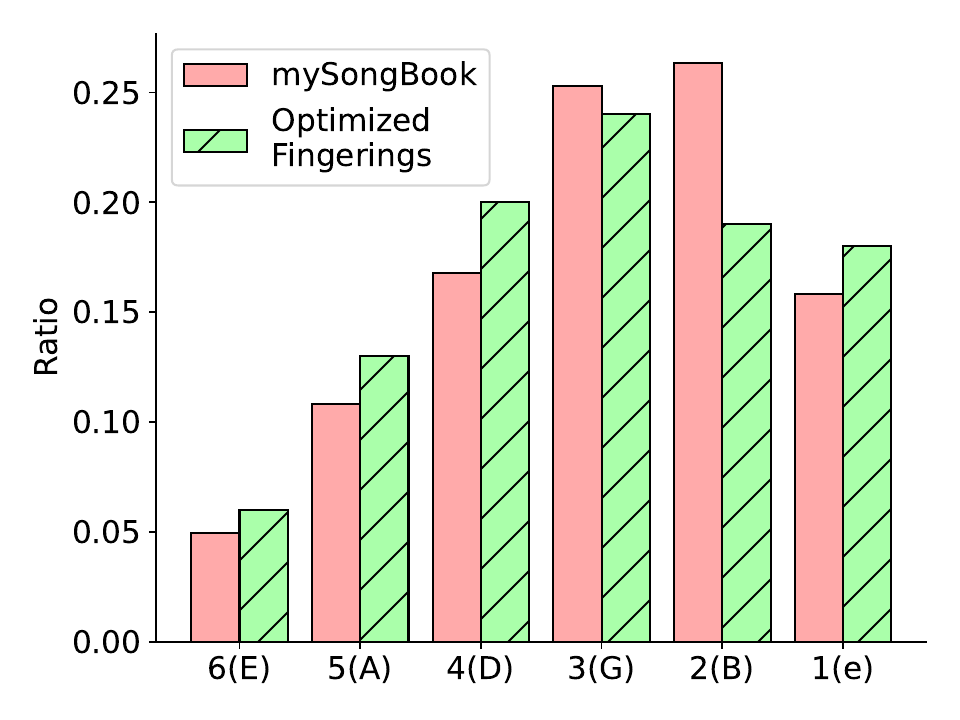}
    \caption{Ratio of the notes played on the different strings to the total number of notes in the test dataset (applying our optimized fingering) and the entire \textit{mySongBook} corpus.}
    \label{fig:notes-distrib}
\end{figure}

The data were also used as a comparison benchmark for the basic fingerings generated by the optimization procedure. On a test dataset made up of lead parts comprising a total of around 1500 notes and by setting the configuration weights in such a way as to balance the different discomfort factors, it was observed that the distribution of notes for each string was reasonably similar to that found in the entire \textit{mySongBook} corpus (\autoref{fig:notes-distrib}).

Finally, specifically regarding bending techniques, a manual analysis of numerous transcriptions taken from the \textit{mySongBook} corpus and from teaching manuals \cite{Fischer_1995_masters_rock, Stetina_1992_metal, Wheatcroft_2008_blues} revealed some typical situations where bends are used, specifically in the presence of two or more consecutive notes in unison, or of groups of three consecutive notes with first and third notes equal to each other and the second one with higher or lower pitch within the limits consonant with the technique in question (e.g., G4-A4-G4, where A4 is performed in bending, keeping the fret/chord/finger combination steady), or when two contiguous notes can be played at the same fret of the same string, using a bend on one of them. 
Such typical situations are again used as a rule-of-thumb for bending decisions, as described in the next Section.

\subsection{Inclusion of articulations and expressive techniques}
\label{sec:articulations-inclusion}

The dataframe generated from the MIDI melody, combined with the fingering solution obtained from the optimization procedure, is further processed in Python to insert articulations and expressive techniques. 
Those considered are listed below, along with a summary of the rules for evaluating the possibility or opportunity of their execution:
\begin{itemize}
    \item hammer-on: only the parameter of biomechanical feasibility is evaluated. If the note is preceded by another lower-pitched note on an open string or played with a lower finger on the same string, the articulation is possible;
    \item pull-off: only the parameter of biomechanical feasibility is evaluated. If the note is preceded by another higher-pitched note played with a higher finger on the same string, the articulation is possible. The pull-off is always possible between fretted and open strings;
    \item vibrato: two user-customizable duration values are set; if the note duration reaches the first, the expressive technique is considered possible; if it reaches the second it is estimated highly likely. Vibrato is not allowed on open strings;
    \item slide legato: only the parameter of biomechanical feasibility is evaluated. If a note is followed by another note on the same string and at a different fret, played with the same fretting hand finger, the articulation is possible;
    \item bending: the field of bending is highly articulated and complex to model \cite{DHooge_2023_bends}. For this first implementation of the software, only basic upward bends (and possibly the releases associated with them, if the following note is played with the same finger and at the same string/fret combination) were considered. The user is first asked to set the maximum amplitude in semitones of the bends that can be performed by each finger of the fretting hand. Setting this parameter to 0 excludes fingers (typically the little finger) from the possibility of doing bends. Bending is considered possible if the hypothetical application note has a duration equal to or greater than a user-settable value. Biomechanical compatibility between the execution of the bending and the fingering of the previous and next notes is also taken into consideration. The procedure tries to include bends by recreating where possible the cliché situations described in Section \ref{sec:stats}.
\end{itemize}

The actual distribution of articulations or expressive techniques depends on the user's choices and the reference statistics described in Section \ref{sec:stats}.
The user can let such statistics 
be taken as targets to be achieved, consistent with the biomechanical possibility of the execution. In other words, the procedure tries to include, where possible according to the conditions described above, the techniques to respect the proportions between technique use and the total number of notes for each string. If the points of possible insertion of the techniques are insufficient in number to reach the targets, the procedure will simply insert the techniques wherever possible, without forcing their insertion into unnatural contexts. If, on the other hand, the potential insertion points are more numerous than the target, the use of the different techniques is chosen on a random basis, to get as close as possible in the overall part to the target values. In the case of vibrato, priority is given to the longest notes. 

The user also has the possibility to customize the reference targets in relation to each of the above-mentioned techniques, so as to be able to better adapt the system's response according to musical genre, personal preferences, or educational needs. The user can also ask the procedure to insert vibratos, hammer-ons, and pull-offs wherever possible.
Only in the case of the slide legato, given the small number of potential occurrences, it was deemed appropriate to limit ourselves to asking the user if s/he would like different notes played with the same finger on the same string to be connected with this technique.

We remark that, clearly, the system also takes into consideration the compatibility between the different techniques, so as not to overlap conflicting ones.

    \section{MusicXML conversion and graphical rendering}\label{sec:MusicXML}

Once the dataframe is completed, including all useful information concerning pitch, timing, use of open or fingered strings, articulations, and expressive techniques, a file in MusicXML format is generated via a dedicated Python script. The MusicXML format is considered a standard for the interchange and distribution of musical scores, provides for tablature encoding, and is supported by many music notation programs \cite{Good_2001_musicxml, Good_2003_XML-interchange}. We opted for \textit{Guitar Pro 8}, to obtain a graphical rendering of the processed parts. 
\begin{figure*}[ht!]
    \centering
    \includegraphics[width=0.85\textwidth]{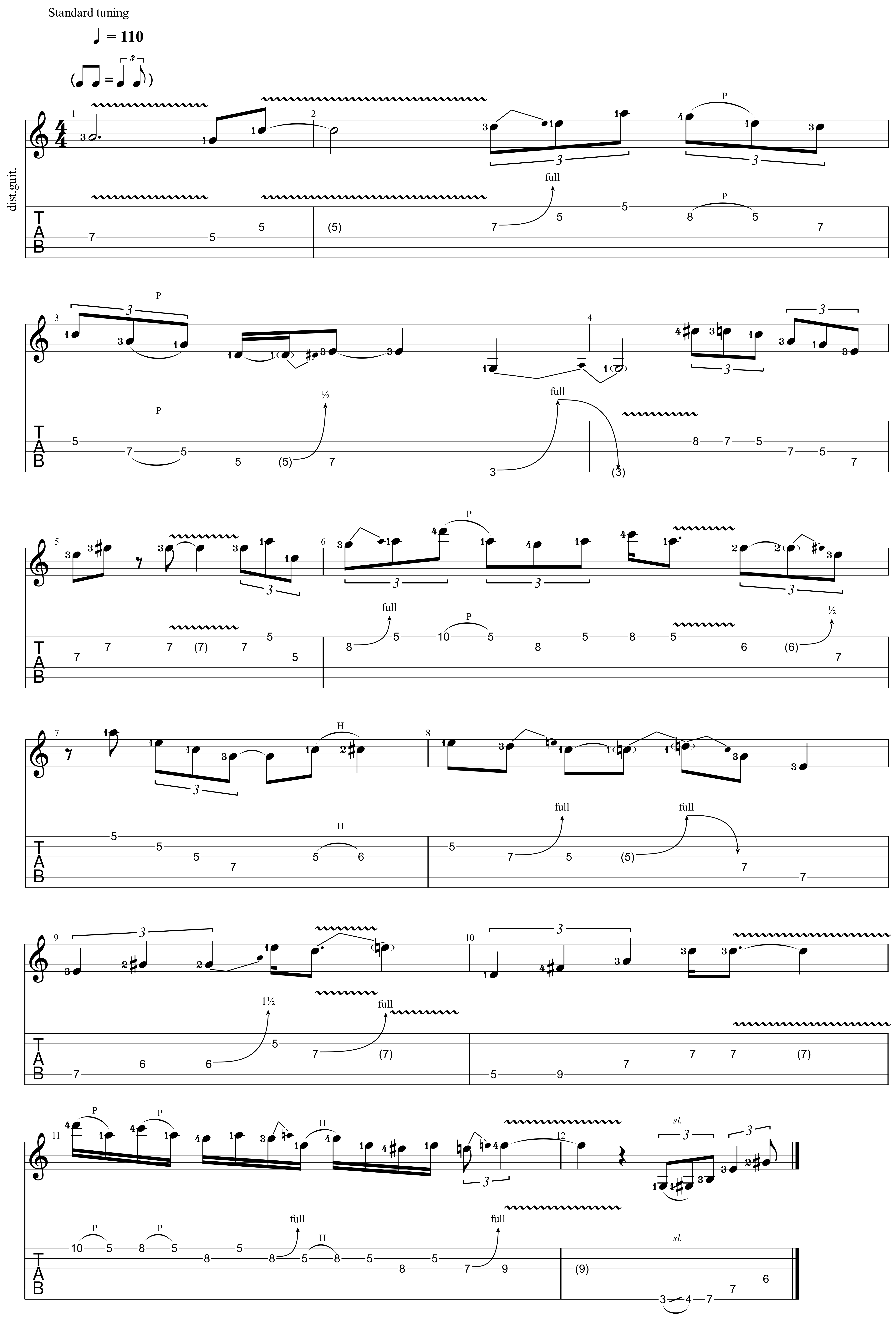}
    \caption{Rich tablature and standard notation of a sample lead part automatically generated by our system.\label{fig:lead-example}}
\end{figure*}

The final result concerning a sample lead part, processed by the entire system, is shown in Figure \ref{fig:lead-example}. The melody is an original work composed by the first author, but any MIDI melody could be used. The part was specifically written by imagining a standard 12-bar blues harmonic progression in A. Here, the reader may notice that the fingering is biomechanically valid and comfortable. The rich presence of bends, vibratos, hammer-ons, and pull-offs is because higher target values were given when inserting articulations and expressive techniques than those found in the \textit{mySongBook} corpus, to better embody the blues or rock/blues feel of the phrasing \cite{Wheatcroft_2008_blues}. 
A slide legato also appears in the last bar.

\section{Conclusions}
\label{sec:conclusions}

We presented an innovative automatic system for generating (starting from simple MIDI melodies) lead guitar tablatures, enriched with fingering as well as articulation and expressive techniques.	 
For the basic fingering, we used an optimization approach significantly richer than the ones already existing in the literature, that allowed us to embed more complex requirements. 
Moreover, we also developed a procedure that provides a very high level of customization for the user, both in terms of biomechanical preferences/requirements and stylistic intents. Starting from the basic fingering, a user-customizable rule-based system, which evaluates note by note the biomechanical feasibility and probability of use of specific articulations and expressive techniques, introduces such techniques whenever appropriate.
Finally, the generated dataframe is converted into MusicXML format, compatible with numerous music notation software. The code is available upon request to the authors.

The computational generation of optimal or alternative \textit{rich} fingerings for a given melodic part can be of interest, beyond the support of the performer in the study phase, in multiple areas such as teaching \cite{Izumi_2002_learn}, automatic music generation \cite{Carnovalini_2020_creativity, Civit_2022_review}, computational performative expressiveness \cite{Cancino-Chacón_2018_review, Bontempi_2023_popular-expressiveness}, stylistic analysis, evaluation of existing tablatures \cite{Radisavljevic_2004_path}, and testing of audio-to-tablature models. Both the Centro di Sonologia Computazionale of the University of Padua \cite{Canazza_2020_CSC} and the TABASCO project team (\url{http://algomus.fr/tabasco/}, last accessed: 2024/05/16), to which part of the authors belong, share previous relevant work and scientific interest in relation to these issues. We believe that in each of the fields mentioned above the verification of performative plausibility and the possible generation of alternative \textit{rich} fingerings can prove valuable (if not indispensable) aids. 


The future improvements of the system's functionalities may comprise the management of lead parts including multiple simultaneous notes, the possibility of inserting decisions concerning articulations and expressive techniques within the optimization procedure, as well as the application of the system in different fields such as teaching (even self-taught) and computational musical expressiveness. Moreover, the much-needed integration of a system such as the one proposed here into the workflow of contemporary music production, often based on the use of virtual instruments, would allow even those who are not expert guitarists to obtain more realistic and convincing parts.
Finally, 
it should be noted how a system such as the one proposed cannot be easily validated, as a given starting melody may correspond to many divergent solutions, equally valid but responding to different expressive intentions and personal preferences. In other words, it would not be correct to compare the output of the system to specific actual performances or existing transcriptions to sanction or not its validity. Although the results obtained so far have been evaluated by the authors as fully satisfactory and realistic, it would therefore be desirable in the future to subject them to the judgment of a wider audience of experienced guitarists.

	\begin{acknowledgments}
	This work is partially supported by the French National Research Agency, in the framework of the project TABASCO (ANR-22-CE38-0001).\ The authors would also like to thank \textit{Arobas Music} for providing the dataset described in Section \ref{sec:stats}.
	\end{acknowledgments} 
	
	\bibliography{smc2024bib}
	
\end{document}